\documentclass[11pt,a4paper]{article}
\usepackage[hyperref]{naaclhlt2019}
\usepackage{times}
\usepackage{latexsym}

\usepackage{amsmath}
\usepackage{amssymb}
\usepackage{subfigure}
\usepackage{multirow}
\usepackage{graphicx}
\usepackage{url}

\aclfinalcopy 

\title{Medical code prediction with multi-view convolution and description-regularized label-dependent attention}

\author{Najmeh Sadoughi$^1$, Greg P. Finley$^1$, James Fone$^1$, Vignesh Murali$^1$, \\
{\bf Maxim Korenevski$^1$,  Slava  Baryshnikov$^1$, Nico }\\
{\bf  Axtmann$^2$, Mark Miller$^1$, David Suendermann-Oeft$^1$}\\
  1. EMR.AI Inc.,  San Francisco, CA, USA \\
    2. DHBW, Karlsruhe, Germany\\
  {\tt najmeh.sadoughi@emr.ai}\\
  }
  
\date{}

\begin{document}
\maketitle
\begin{abstract}
 A ubiquitous task in processing electronic medical data is the assignment of standardized codes representing diagnoses and/or procedures to free-text documents such as medical reports.
This is a difficult natural language processing task that requires parsing long, heterogeneous documents and selecting a set of appropriate codes from tens of thousands of possibilities---many of which have very few positive training samples.
We present a deep learning system that advances the state of the art for the MIMIC-III dataset, achieving a new best micro F1-measure of 55.85\%, significantly outperforming the previous best result \cite{mullenbach2018explainable}.
We achieve this through a number of enhancements, including two major novel contributions:
multi-view convolutional channels, which effectively learn to adjust kernel sizes throughout the input;
and attention regularization, mediated by natural-language code descriptions, which helps overcome sparsity for thousands of uncommon codes.
These and other modifications are selected to address difficulties inherent to both automated coding specifically and deep learning generally.
Finally, we investigate our accuracy results in detail to individually measure the impact of these contributions and point the way towards future algorithmic improvements.
\end{abstract}
\section{Introduction}
Coding medical reports is the standard method used by health care institutions for summarizing patients' diagnoses and the procedures performed on them. Among other things, medical codes are used for billing, epidemiology assessment, cohort identification, and quality control of health care providers.

Assignment of standardized codes, though valuable, is a difficult task even for human coders.
Part of this challenge is in the sheer number of codes:
the United States version of the Ninth Revision of the International Classification of Diseases (ICD-9-CM), for instance, contains approximately 18,000 procedure and diagnosis codes; the current 10th Revision (ICD-10) includes even more codes, approximately 171,000.\footnote{\url{https://www.cdc.gov/nchs/icd/icd10cm_pcs_background.htm}}
Additionally, the amount of data to be processed is substantial: coding inpatient charts typically involves reviewing multiple notes such as discharge summaries, progress notes, operative notes, and physician, nurse, or attendee notes; any of these notes may individually evidence for specificity of one or more codes.
Code sets are often subject to annual revision, making constant re-training and feedback to coders necessary.
Compounding the task's general difficulty, there is a degree of subjectivity in coding which can result in discrepancies even between well-trained, highly accurate coders \cite{farkas2008automatic}.
All of these factors contribute to increased errors by human coders.

Automated coding or AI-assisted coding approaches can reduce the time and effort spent by humans for annotating the reports and, even more importantly, potentially reduce their errors.
Given enough annotated data, machine learning algorithms trained on data annotated by multiple coders can  dilute the subjectivity of individual judgments, and hence reduce subjectivity error \cite{farkas2008automatic}.

However, automated coding also shares many of the aforementioned challenges. Moreover, the non-uniformity of distributions of diseases and procedures results in large number of sparse classes, for which very few positive training cases are available. Data sparsity can also be a problem when code set revisions introduce new codes for which no annotated data is initially available.
Hence, there is a need for machine learning approaches which are generally more robust to data sparsity. 
	
We propose a new end-to-end neural network model to solve the prediction of codes from medical reports as a multi-task classification problem, achieving a new state-of-the-art result on the MIMIC-III corpus, which is the largest publicly available dataset for this task. Our model benefits from two novel contributions, namely multi-view CNN channels and label-dependent attention layers tuned to label descriptions. That is, our model exploits the description of the codes for regularizing the attention for each individual classifier, reducing the effect of data sparsity. We also demonstrate the benefit of using all notes, in contrast to only using the discharge summaries as in previous studies. 

\section{Related work}
There has been significant work towards the automated coding problem \cite{perotte2013diagnosis,kavuluru2015empirical,wang2016diagnosis,scheurwegs2017selecting,prakash2017condensed,rajkomar2018scalable,amoia2018scalable}. We review some of the recent relevant work.

\citet{perotte2013diagnosis} relied on the MIMIC-II dataset and experimented with flat and hierarchical support vector machines (SVMs) on tf-idf features. For hierarchical SVMs, they exploit the knowledge of ICD-9 code hierarchies. They demonstrated that using hierarchical SVMs increased the recall for sparse classes and achieved superior performance compared to flat SVMs.

\citet{kavuluru2015empirical} built classifiers for ICD-9 diagnosis codes over three datasets, the biggest of which included around 71K EMR discharge summaries and 1,231 distinct codes. They performed feature selection and used a variety of methods such as SVM, na\"ive Bayes, and logistic regression for this problem. Using an ensemble of these classifiers, they achieved a micro F1-score of 0.479 on their biggest corpus.

\citet{baumel2018multi} used the publicly available MIMIC-III and MIMIC-II datasets and proposed new deep models including a CNN model and a hierarchical GRU model with label-dependent attention layer for ICD-9 diagnosis code prediction.  Their best performance on MIMIC-III was achieved by a CNN model, obtaining a micro F1-score of 40.7\% for diagnosis codes.

\citet{mullenbach2018explainable} presented a model capable of predicting full codes for both ICD-9 diagnoses and procedures composed of shared embedding and CNN layers between all codes and an individual attention layer for each code. They also proposed adding regularization to this model using code descriptions. Their best model on MIMIC-III reached a micro F1-score of  53.9\%, which was achieved by their base model without regularization.

\citet{wang2018joint} proposed a model which jointly captures the words and the label embeddings and exploits the cosine similarity between them in predicting the labels. They applied this model to the task of predicting only the most frequent 50 codes in MIMIC-III, which they accomplished with a micro F1-score of 61.9\%.

Our approach is most similar to the current state-of-the-art model by \citet{mullenbach2018explainable}.
As in their study, we use a CNN layer with attention modules by code and approach regularization using code descriptions.
Our model has notable departures from theirs, however: firstly, we use multi-view CNN channels with max pooling across the channels, which in itself leads to improvements over their model (even before attention regularization). Secondly, they did not demonstrate any improvements by using code descriptions in regularizing their model on MIMIC-III, when predicting full codes. Whereas they regularize the last layer,  we regularize the attention layer, leading to improvements over our base model. Our use of independent attention layer for each code is also similar to the approach used by \citet{baumel2018multi}, where they used shared GRU layers over the sentences across the labels and then performed label-dependent attention pooling for each class. However, their model is RNN based and ours is CNN based, and they did not achieve superior performance for this model compared to a fully shared CNN model with max pooling  when predicting all diagnosis codes in MIMIC-III.

\section{Database}
\label{sec:data}
We rely on the publicly available MIMIC-III dataset \cite{johnson2016mimic} for ICD-9 code predictions.\footnote{Note that some previous studies reported results on MIMIC-II, which is an older version and a subset of MIMIC-III. Due to space limitations, and the fact that MIMIC-III is more comprehensive and current, we only focus our experimental efforts on MIMIC-III.} This dataset includes the electronic medical records (EMR) of inpatient stays in a hospital critical care unit. MIMIC-III includes raw notes for each hospital stay in different categories---\emph{discharge summary report}, \emph{discharge summary addendum}, \emph{radiology note}, \emph{nursing notes}, etc. The number of notes varies between different hospital stays. Also, some of the hospital stays do not have discharge summaries; following previous studies for automated coding, we only consider those that do \cite{perotte2013diagnosis,baumel2018multi,mullenbach2018explainable,wang2018joint}.

We have three sets for our experiments: one including only the discharge summaries, which allows us to compare our results with previous studies on this corpus \cite{mullenbach2018explainable,perotte2013diagnosis}, hereon the \emph{Dis} set; one on the concatenation of all patient notes, hereon, \emph{Full} set; and one on another set which includes only discharge summary samples with the 50 most frequent codes, hereon, \emph{Dis-50} set, for comparison to previous studies \cite{mullenbach2018explainable,wang2018joint}.

The dataset includes 8,929 unique ICD codes (2,011 procedures, 6,918 diagnoses) for the patients who have discharge summaries.\footnote{Note that \citet{mullenbach2018explainable} found 8,921 unique codes.
}
We follow the train, test, and development splits publicly shared by the recent study on this dataset \cite{mullenbach2018explainable}. These splits are patient independent. 
The statistical properties of all the sets are shown in Table \ref{tab:stats}; note that there are around three times more tokens for the each hospital admission for the Full set compared to the Dis set. Note that Dis-50 includes far fewer training instances because any instances which do not include any of the 50 most frequent codes are discarded.

For preprocessing the text, we convert all characters to lower case and remove tokens which only include numbers. We build the vocabulary from the training set and consider words occurring in fewer than three training samples as \emph{out of vocabulary} (OOV). This results in 51,917 unique words
for the Dis and Dis-50 set  and 72,891 for the Full set.

\begin{table*}
\centering
\caption{The table demonstrates the properties of the two different sets: Dis and Full. ``hadm'' stands for hospital admissions and code cardinality is the average  number of codes per hospital admission.}
\resizebox{13cm}{!}{
\begin{tabular}{|l |l| rrr| rrr |}
\hline
\multicolumn{2}{|c|}{}&\# hadms&\# tokens & \# types & \# codes & \# unique codes & code cardinality\\
\hline
Dis set& train &47,723&70,846,775&140,796&758,216&8,693& 15.89\\
& test &3,372&6,043,744&42,860&61,579&4,085&18.26\\
& dev &1,631&2,910,871&30,969&28,897&3,012&17.72\\
\hline
Full set& train &47,723&206,783,294&211,891&758,216&8,693 & 15.89\\
& test &3,372&19,266,433&59,436&61,579&4,085& 18.26\\
& dev &1,631&9,209,681&41,786&28,897&3,012&17.72\\\hline
Dis-50 set& train &8,066&12,338,530&59,168&45,906&50&5.69\\
& test &1,729&3,156,603&31,999&10,442&50&6.04\\
& dev &1,574&2,830,896&30,476&925&50& 5.88\\
\hline
\end{tabular}}
\label{tab:stats}
\end{table*}

\section{Method}
\begin{figure*}[t]
\centering
\includegraphics[trim=0cm 0.0cm 0.0cm 0cm, clip, width=14cm]{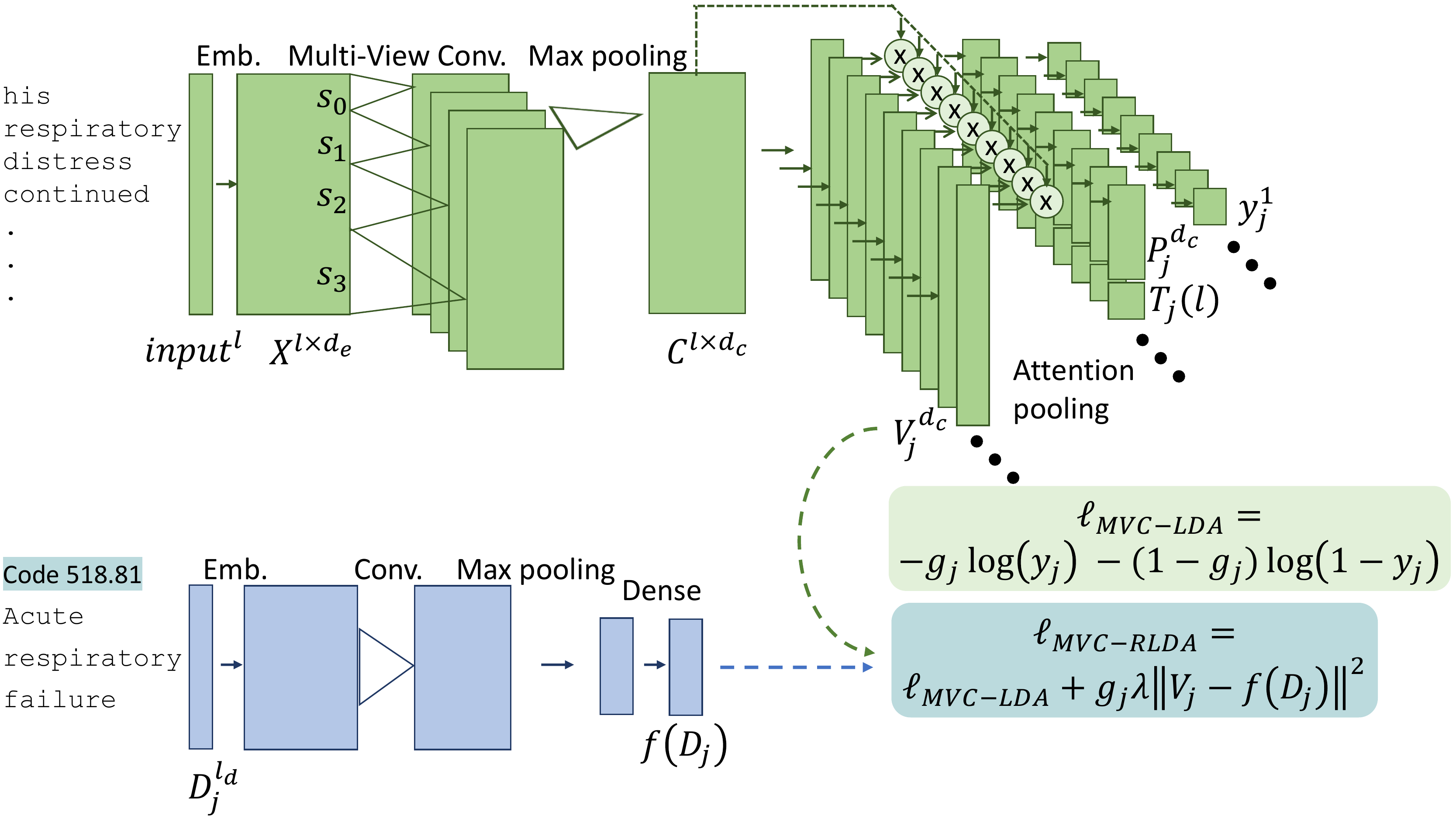}
\caption{The block-diagram of MVC-LDA (green blocks only) and MVC-RLDA (all blocks) models.}
\label{fig:model}
\end{figure*}

We approach the task of predicting ICD codes from medical notes as a multi-task binary classification problem in which each code in each hospital admission can be present (labeled 1) or absent (labeled 0). We build our model with an embedding layer stacked with multi-view CNNs to selectively capture the relationships between a set of n-gram embeddings and ICD codes. We use max pooling across these CNN channels and rely on attention spatial pooling. While the embedding layer and multi-view CNNs are shared between all codes, we consider  individual attentions for different codes. Separately modeling the attention can help in interpreting the predicted labels.
This constitutes our base model, hereon, \emph{multi-view convolution with label-dependent attention pooling} (MVC-LDA). We enhance this model by using the natural-language descriptions of ICD codes to regularize the attention layers during training, enforcing similar codes in description embedding space to have similar attentions. We call this model \emph{multi-view convolution with regularized label-dependent attention pooling} (MVC-RLDA).
The architectures of these models are visualized in Figure \ref{fig:model}.

\subsection{Embedding layer}
\label{sec:emb}
The first layer of our model maps words to their continuous embedding space.		
Each word $w \in \mathbb{R}^{d_v}$ is mapped to $x \in \mathbb{R}^{d_e}$ using the embedding weight $W_e \in \mathbb{R}^{d_v \times d_e}$, where $d_v$ is the vocab size and $d_e$ is the embedding dimensionality. We consider all embedded words from one input note with length $l$ as $\mathcal{X}=\left[x_0, x_1, ..., x_{l-1}\right]^T \in \mathbb{R}^{l\times d_e}$.
Our pilot experiments demonstrated enhancement of the classification results when we used pre-trained embeddings compared to random initialization. 
Hence, we pre-train the embedding layer on all text in the training set using the Gensim implementation of the \emph{continuous bag-of-words} (CBOW) word2vec approach \cite{mikolov2013efficient,mikolov2013linguistic}, with an embedding size of 100, trained over a window size of 5, with no minimum count for 5 epochs.

\subsection{Multi-view convolutional layer}
Our goal behind using multiple channels with different kernel sizes is that the underlying  informative n-gram   in the input for each code can vary in length according to the word neighborhood, and using multiple different field views within a CNN has the potential to capture that. 

Our multi-view convolutional layer consists of 4 convolutional channels with different kernel sizes ($s$, $s-2$, $s-4$ and $s-6$, where $s$ is the biggest kernel size), and the same number of filters with stride of 1 (Fig. \ref{fig:model}). 
To preserve the input length, we perform zero-padding on the input to this layer, $\mathcal{X}$. We apply  max pooling across the outputs of these four channels. For the $n^{th}$ word of the input, and assuming an odd kernel size for the $i^{th}$ channel, Equation \ref{eq:conv} calculates the output of this layer, where $s_i$ is the kernel size, and $W_i \in \mathbb{R}^{s_i\times d_e\times d_c}$ is the convolution weight. After cross-channel max pooling for the whole input, the output of this layer is $\mathcal{C} = [c_0, c_1, ..., c_{l-1}] \in \mathbb{R}^{l \times d_c}$. This convolutional layer is shared across classes, and therefore is assumed to capture the relevant n-grams for all of them.
Note that multi view CNNs have been used before by \citet{kim2014convolutional} for sentence classification. However, \citeauthor{kim2014convolutional} used spatial max pooling over the CNN channels and concatenated them, whereas we use max pooling across the channels to select the most relevant n-gram for each filter. Therefore, our method flexibly picks the most salient channels according to the input. 
\begin{equation}
c_n = \mathrm{tanh}\left(\underset{i \in [0,3]}{\mathrm{max}} W_i \ x_{n-\frac{s_i}{2}:n+\frac{s_i}{2}}\right)
\label{eq:conv}
\end{equation}

\subsection{Attention layer}
For spatial pooling of the convolutional outputs, we rely on an attention mechanism. We consider separate attention layers for each class of output (Fig. \ref{fig:model}). Since there is a large number of output classes (8,929), this helps the model in attending to relevant parts of input for each output separately. For modeling the attention for each class, we use a linear layer with weight $V_j \in \mathbb{R}^{d_c}$ for the $j^{th}$ class. The attention for the input $\mathcal{C}$ and class $j$ is calculated by $\mathcal{C}V_j$.
The derived attentions are used to weight each frame from the convolutional layer output before pooling them. Equation \ref{eq:attn} shows how this function is performed, where $P_j \in \mathbb{R}^{d_c}$ is the pooled output (see attention pooling in Figure \ref{fig:model}).
\begin{equation}
P_j = \mathcal{C}^T  \left(\mathcal{C} \ V_j \right)
\label{eq:attn}
\end{equation}

\subsection{Output layer}
The output for each class is a dense layer with sigmoid nonlinearity. During testing, output values greater than 0.5 are assigned as present (1) and the rest are assigned as absent (0). 
Note that since for each class the majority of the training samples have output of 0 (e.g., even the most frequent code in the training examples occurs in only 37\% of training instances), the network is strongly biased towards negative predictions. In our preliminary experiments, we found that the network usually tends to under-code---i.e., the cardinality of predictions were lower than the ground truth. 
We found a positive and statistically significant Pearson's correlations between the input length and the number of ground truth codes (for training samples in Dis set:
$\rho=0.479$, $p<.001$, and in Full set: $\rho=0.557$, $p<.001$).	
Therefore, we use the input length as an extra conditioning input to the output layer to shift the bias of the output sigmoid layer from zero accordingly, to cope with the problem of under-coding to some extent. 

We embed the input length using Equation \ref{eq:len}, where $T_j$ is the embedding function for the $j^{th}$ class, $l$ is the input length for an arbitrary sample, $K_j \in \mathbb{R}$ is the layer weight, and $d_j \in \mathbb{R}$ is its bias. Note that the under-coding may differ from one class to another due to the difference in their occurrence frequencies, hence we use separate length embedding functions  for each  class to capture these differences. Moreover, using a non-linear function such as sigmoid in the embedding function has more flexibility and helps the model to generalize better to unseen input lengths in the two extremes.
\begin{equation}
T_{j}(l)=\mathrm{Sigmoid} \left(K_jl + d_j\right)
\label{eq:len}
\end{equation}
The embedded length is incorporated in the output layer as shown in Equation \ref{eq:out}, where $U_j \in \mathbb{R}^{d_c}$ is the weight and $b_j \in \mathbb{R}$ is the bias, and $y_j \in \mathbb{R}$ is the prediction for the $j^{th}$ class. Note that $T_{j}(l)$ shifts the bias of the output sigmoid layer according to the input length (Fig. \ref{fig:model}).
\begin{equation}
y_j = \mathrm{Sigmoid} \left(U^T_j \ P_j + b_j + T_{j}(l)\right)
\label{eq:out}
\end{equation}

We use the binary cross entropy loss function on the output of this layer as shown in Equation \ref{eq:loss1}, where $g_j$ is the ground truth for the $j^{th}$ class. For each batch, the loss function is calculated for each sample and is averaged across them. MVC-LDA is trained to minimize this loss function. The top blocks in Figure \ref{fig:model} summarize the entire MVC-LDA model.
\begin{equation}
\ell_{MVC-LDA} = -g_j\mathrm{log}\left(y_j\right) - (1 -g_j)\mathrm{log}\left(1-y_j\right)
\label{eq:loss1}
\end{equation}

\subsection{Regularizing attention by label description}
As mentioned earlier, many classes are quite rare in the data, so their attention modules are trained on very few examples.
To better handle these cases of sparsity, we relied also on the label descriptions included in MIMIC-III (e.g., 518.81: `Acute respiratory failure'; 37.22: `Left heart cardiac catheterization').
We hypothesized that a code's description is semantically and lexically similar to the segments of input text that contain positive evidence for that code.
Thus, we devised a means of directing attention via regularization, constraining the attention weight $V_j$ for the $j^{th}$ class by its description.

We map the description of labels to an embedding space using a nonlinear function  $f$, a neural network composed of an embedding layer tied with $W_e$, a convolutional layer with kernel size of $s$, and $d_c$ filters, a spatial max pooling, and a nonlinear dense output layer with sigmoid function (See the blue blocks in  Figure \ref{fig:model}).

Suppose the description of the $j^{th}$  label is $\mathcal{D}_j \in \mathbb{R}^{l_w}$. During training, whenever the gold standard contains the $j^{th}$ code, we add a regularization term to the loss function in Equation \ref{eq:loss1}, resulting in the loss function shown in Equation \ref{eq:loss}, where $g_j$ is the ground truth label for the $j^{th}$ class and $\lambda$ specifies the weight of the regularization  in the new loss function. Adding this extra term in the loss function constrains the training of the attention weights to avoid overfitting, particularly for classes with few training samples, by pushing the attention weights to be closer to the description embeddings for each class. Moreover, this regularization indirectly pushes the attention for classes with similar descriptions to be closer to each other.
\begin{equation}
\ell_{MVC-RLDA} = \ell_{MVC-LDA} + g_j \lambda \| V_j -f\left(\mathcal{D}_j\right)\| ^2 
\label{eq:loss}
\end{equation}

\section{Experiments}
\subsection{Baselines}
We compare our approach with four baselines: flat and hierarchical SVMs \cite{perotte2013diagnosis}, LEAM \cite{wang2018joint}, and CAML \cite{mullenbach2018explainable}. 

For flat and hierarchical SVMs, we follow the approach of \citet{perotte2013diagnosis}, considering 10,000 tf-idf unigram features, training 8,929 binary SVMs for the flat SVMs and 11,693 binary SVMs for hierarchical SVMs. For the hierarchical SVMs, we use the ICD-9-CM hierarchy from bioportal.\footnote{https://bioportal.bioontology.org/ontologies/ICD9CM} For flat SVMs, a code is considered present if its SVM predicts a positive output. For hierarchical SVMs, a code is considered present if the SVMs for the code and SVMs for the all parents of the code are positive.

LEAM \cite{wang2018joint} learns the joint representation of labels and input embeddings and uses their cosine similarity in predicting the codes. We compare our model with their results on the Dis-50 set.

CAML \cite{mullenbach2018explainable} has achieved the best state-of-the-art results on MIMIC-III. CAML is composed of a stack of an embedding layer, a CNN layer, and label-dependent attention layers. We run their model on the Dis set using their publicly available code.\footnote{https://github.com/jamesmullenbach/caml-mimic} The only difference here is that we found slightly more unique codes: 8,929 to their 8,921.

\subsection{Evaluation metrics}
The most widely used metric for evaluating ICD code prediction is micro F1-score \cite{perotte2013diagnosis,wang2018joint,mullenbach2018explainable,kavuluru2015empirical}. New studies have reported results on macro F1-score, precision@n, and AUC of ROC as well \cite{wang2018joint,mullenbach2018explainable}.
As evaluation metric we rely on  \emph{micro F1-score} (micro F1), \emph{macro F1-score} (macro F1) for the top 50 codes, \emph{area under the precision-recall curve} (PR AUC) and \emph{precision@n} (P@8 when evaluating the models on all codes and P@5 when evaluating the models for the top 50 codes). For obtaining the micro F1, the micro precision and recall are calculated by collapsing all classes into one class and considering the task as a single binary classification. Therefore,  micro F1 weighs all class occurrences similarly. On the other hand, macro precision and recall are calculated by evaluating the precision and recall for each class and then averaging those values across the classes, weighting all classes similarly. However, since the number of ground truth occurrences for 54\% of the codes in Dis and Full sets are zero (see Table \ref{tab:stats}), the recall for those values can not be calculated. 
Therefore, we do not report macro F1 when evaluating the models on all codes. For the Dis-50 set, however, there is no such problem, as all testing codes  occur in training.

Furthermore, depending on the application, one may tune the threshold for binary classification. Hence, we also report PR AUC, which provides the area under the curve of micro recall  versus micro precision for thresholds between 0 and 1. We do not use ROC AUC as a metric, as this measures true negatives, which are extremely frequent for this problem and thus yield very high and uninformative scores.
P@n gives the precision of the $n$ highest prediction scores for each sample and is motivated by assessing the performance of the auto-coder model in an AI-assist workflow, where a human coder would hypothetically be provided with the top $n$ predictions for each note.

\begin{table*}[t]
\centering
\caption{Hyperparameter value candidates searched using Hyperband, and their optimal values.}
\resizebox{11cm}{!}{
\begin{tabular}{|l|l|l|l|l|}
\hline
model &data& hyper parameters & candidates & selected\\
\hline
\multirow{6}{*}{MVC-LDA}  & \multirow{2}{*}{Dis-50} & $(s_0, s_1, s_2, s_3)$&(2-8, 4-10, 6-12, 8-14)& (2, 4, 6, 8)\\
  &  & $d_c$&30-100& 90 \\
\cline{2-5}
& \multirow{2}{*}{Dis} & $(s_0, s_1, s_2, s_3)$&(2-8, 4-10, 6-12, 8-14)& (6, 8, 10, 12)\\
 &  & $d_c$&30-100& 70\\
 \cline{2-5}
 & \multirow{2}{*}{Full} & $(s_0, s_1, s_2, s_3)$&(2-8, 4-10, 6-12, 8-14)& (8, 10, 12, 14)\\
 &  & $d_c$&30-100& 90 \\
 \hline\hline 

\multirow{9}{*}{MVC-RLDA} & \multirow{3}{*}{Dis-50} & $(s_0, s_1, s_2, s_3)$&(2-8, 4-10, 6-12, 8-14)& (6, 8, 10, 12)\\
 &  & $d_c$&30-100& 90\\
 & & $\lambda$& \{0.001, 0.0001, 0.005, 0.05, 0.01\} & 0.005\\ 
 \cline{2-5}
 & \multirow{3}{*}{Dis} & $(s_0, s_1, s_2, s_3)$&(2-8, 4-10, 6-12, 8-14)& (6, 8, 10, 12)\\
 &  & $d_c$&30-100& 90\\
 & & $\lambda$& \{0.001, 0.0001, 0.0005, 0.0007, 0.01\} & 0.0005\\
 \cline{2-5}
 & \multirow{3}{*}{Full} & $(s_0, s_1, s_2, s_3)$&(2-8, 4-10, 6-12, 8-14)& (6, 8, 10, 12)\\
 &  & $d_c$&30-100& 90\\
 & & $\lambda$& \{0.001, 0.0001, 0.0005, 0.0007, 0.01\} & 0.0005\\

\hline
\end{tabular}}
\label{tab:hyp}
\end{table*}

\begin{table*}
\centering
\caption{Comparisons of our models with baselines. All values reported in percentage format ([\%]).}\resizebox{16cm}{!}{
\begin{tabular}{|l|l  ||ccc| cc||cccc|}
\hline
\multirow{3}{*}{set} & \multirow{3}{*}{method}& \multicolumn{5}{|c||}{all codes} & \multicolumn{4}{c|}{ 50 most freq. codes}\\
\cline{3-11}
&  & \multicolumn{3}{c|}{micro F1} &
\multirow{2}{*}{P@8}&\multirow{2}{*}{ PR AUC  }&\multirow{2}{*}{micro F1} & \multirow{2}{*}{macro F1}&\multirow{2}{*}{P@5}& \multirow{2}{*}{PR AUC}\\
& & Proc. & Diag. & && &&&&\\
\hline
\multirow{4}{*}{Dis-50}&LEAM \cite{wang2018joint}&-&-&- &- &-&61.9& 54.0&61.2&-\\
&DR-CAML \cite{mullenbach2018explainable}&-&-&- &- &-&63.3&57.6&61.8&- \\
\cline{2-11}
&MVC-LDA &-&-&- & -&-&66.82& 59.65&64.43&73.42\\
&MVC-RLDA &-&-&- & -&-&67.41& 61.47&64.11&71.76\\
\hline\hline
\multirow{5}{*}{Dis}&flat SVMs &42.12&38.50& 39.67
& -&-&60.25&52.39&-&- \\
&hierarchical SVMs &45.97&43.25&44.13 & 
-&- &61.43&55.28&-&-\\
&CAML  \cite{mullenbach2018explainable}
&59.05&49.86& 52.03& 
69.48&54.85
&70.21&63.83&66.33&74.12 \\
\cline{2-11}
 &MVC-LDA &61.17&52.25&54.27&
 70.50&56.28&
 72.69&67.39&67.77&76.73\\
 &MVC-RLDA &62.08&52.97&54.96&71.57&\textbf{56.70}&
 72.54&67.86&68.00&77.37\\
 \hline
 \hline
\multirow{2}{*}{Full} &MVC-LDA &63.00&52.65&54.92&
70.58&56.40&
73.13&67.85&68.06&77.95\\
 &MVC-RLDA &\textbf{63.96}&\textbf{53.53}&\textbf{55.85}&
\textbf{ 72.08}&56.19&
\textbf{73.80}&\textbf{68.65}&\textbf{68.56}&\textbf{78.47}\\
\hline
\end{tabular}}
\label{tab:res}
\end{table*}

\subsection{Experimental details}
We used PyTorch for building and training our models.
We train our base model (MVC-LDA) and our regularized model (MVC-RLDA)  for the three sets (i.e., Dis-50, Dis and Full). 
As our optimizer, we rely on Adam with learning rate of $0.001$ and $\beta_1=0.9$, $\beta_2=0.999$, $\epsilon=1e-8$. To optimize other hyperparameters, we use Hyperband \cite{li2017hyperband}, an algorithm for expediting random search on hyperparameters for machine learning models, making it 5 to 30 times faster than Bayesian optimization. Hyperband requires specifying a \emph{resource} to maximally exploit to find the best parameters; we set this to 27 training epochs, as well as an additional pruning parameter $\eta$, which we set to 3. We chose to optimize three hyperparameters with Hyperband: the number of CNN filters, multi-view CNN kernel sizes, and regularization weight for MVC-RLDA. Table \ref{tab:hyp} shows the hyperparameters we tried for our models and the selected value for each, optimized to maximize the micro F1 on the development set.

During training, we use a batch size of 4 samples, and if the sample length is higher than 10,000, we randomly select a segment with 10,000 words from the  input. During testing, we use the whole input. We train all models with early stopping, using micro F1 on the development set as stopping criterion with a patience of 10 epochs.
We use a dropout of 0.2 in our models for reducing the chance of overfitting,
with a pseudorandom seed before starting the experiments.

\subsection{Evaluation results}
We evaluate the baseline models on the Dis set and Dis-50 sets and provide micro F1 scores for diagnosis and procedure codes for comparability with previous studies \cite{mullenbach2018explainable,perotte2013diagnosis}.

Table \ref{tab:res} provides the evaluation of our models (MVC-LDA and MVC-RLDA) and the four baselines.\footnote{The authors of CAML \cite{mullenbach2018explainable} reported micro F1-Proc=60.9\%,  micro F1-Diag=52.4\%, micro F1=53.9\%, P@8=70.9\% and P@15=56.1\% on the Dis set. However, these results are for 8,921 codes.} When the models are trained and tested on Dis-50 set, our models outperform the previous studies in terms of micro and macro F1, P@5, and PR AUC. This is due to the architecture differences, such as the use of multi-view CNN and better use of description by our model compared to the two baselines. When the models are trained on all codes (Dis or Full sets), we have provided evaluations for all codes and  also the test of 50 most frequent codes (i.e., the test set of Dis-50 and their corresponding Full sets). The results demonstrate that the neural network models trained on Dis or Full sets outperform the models trained on Dis-50 set when evaluated on the top 50 codes in terms of micro and macro F1 and PR AUC.
These improvements may have at least two explanations: one, that the Dis-50 set includes less data, as any documents with no occurrences of any of the top 50 codes are not included; and two, that the larger models jointly learn to predict all codes within a single architecture, which can be thought of as a data-dependent regularization for the top 50 codes. 
The flat and hierarchical SVMs performance are lower than all models on the top 50 codes and on all codes. They use 10K uni-gram tf-idf features, and hence, they are subject to the typical limitations of bag-of-words features (no phrases, syntax, locality, etc.). 
Hierarchical SVMs use the hierarchy of the codes in a form of a tree and utilize the dependency between the codes during training. For each split of the tree, the parent SVM is trained by only the data for its children. This increases the recall for the sparse classes as demonstrated by \citet{perotte2013diagnosis}. Therefore the hierarchical SVMs outperform flat SVMs, but their performance is still lower than the CNN-based models (i.e., CAML, MVC-LDA, and MVC-RLDA). 
CAML, previously the best state-of-the-art model, outperforms both SVM based models in  all metrics; our base model, MVC-LDA, outperforms CAML across the board. This shows that the added multi-view CNN and length of text to our model are helpful. 
On the Dis set, MVC-RLDA outperforms MVC-LDA, achieving the best performance in terms of all metrics except micro F1 on top 50. This may be explained by the fact that regularization by definition of codes is more helpful for sparse classes. Therefore the top frequent codes may not benefit as much from this added feature as the sparse codes (further analysis is provided in the following sections). 
Models trained on the Full set outperform their counterparts on the Dis set, with the overall best performance across all metrics achieved by MVC-RLDA on the Full set, except for PR AUC. This shows that the added notes for each patient may have information which may not be present only in the discharge summaries and therefore are useful in learning the codes.

\begin{figure}[t]
\centering
\includegraphics[width=8cm]{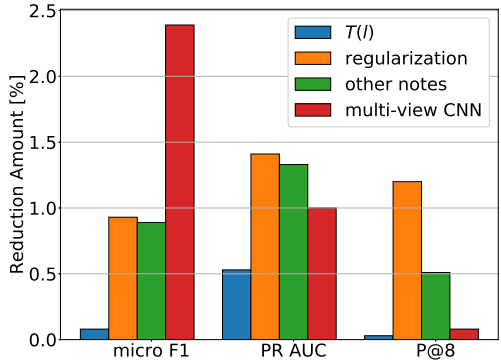}
\caption{The ablation study results: the effect of removing four components (length embedding $T(l)$, regularization, other notes, and multi-view CNN) on micro F1, PR AUC and P@8.}
\label{fig:ablation}
\end{figure}

\subsection{Ablation study}
We perform an ablation study on our best model,
removing certain components one at a time to gauge their respective contributions.
The components we study here are regularization, multi-view CNN, the use of notes other than discharge summaries and  conditioning the output layer on input length embedding ($T\left(l\right)$).
(For most components, we simply remove them individually; for the multi-view CNN, we replace the layer with a simple CNN with a kernel size of 12, which is the maximum kernel size in MVC-RLDA.)

Figure \ref{fig:ablation} shows the the reduction amount of micro F1, PR AUC, and P@8 by removing each of the four components.
The first observation is that removing each component reduces all metrics,  showing their importance to our best model.
Comparing the components with each other shows that length embedding has the lowest effect.
The use of regularization and a reliance on all available notes, on the other hand, are consistently beneficial.
The situation with multi-view CNN is more complex: for micro F1, it has the highest impact, while it has somewhat less on PR AUC and virtually none on P@8.
This difference can be explained by examining precision and recall: removing multi-view CNN decreases recall from 50.17\% to 43.78\%, although precision \emph{increases} (62.97\% to 68.63\%).

PR AUC measures the overall performance of the model across different prediction thresholds; with higher precision and lower recall, the F1-optimal threshold for this model is actually a little lower than 0.5, hence the relatively poor performance at 0.5.

\subsection{Effect of regularization on different labels}
\label{sec:regana}
\begin{figure}[t]
\centering
\subfigure[Micro F1 by MVC-RLDA]{
\includegraphics[width=8cm]{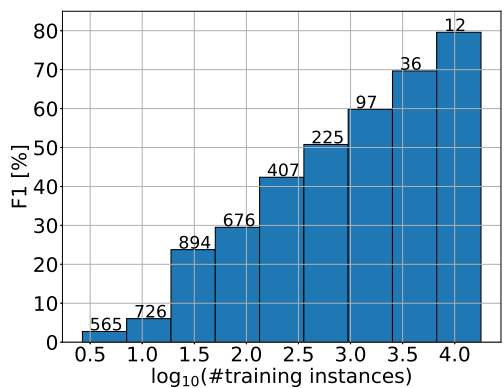}
\label{fig:F1}
}
\subfigure[Improvement of micro F1 after regularization]{
\includegraphics[trim=0cm 15mm 0cm 0cm, clip, width=8cm]{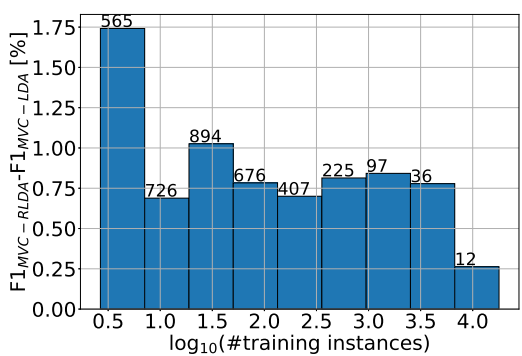}
\label{fig:freqF1}
}
\caption{The effect of regularization with respect to the training support. The number on top of each bar is the number of unique codes (in the test set) for that bar.}
\label{fig:evalfreq}
\end{figure}

In this section we examine micro F1 across different codes in terms of their occurrence frequency in the training set. We limit this analysis to the codes which were both in the training set and in the testing set (i.e., 3956 codes). We divide the range of the logarithm of number of available training instances for the codes into 10 bins. Figure \ref{fig:F1} shows the micro F1-score for the codes across different number of training samples. 
On top of each bar, we have provided the number of codes which belong to that bar in the testing set. This figure clearly shows that micro F1 improves when the number of training samples increases. To see the effect of regularization across the number of available training samples, we plot the difference of the micro F1 achieved by MVC-RLDA from MVC-LDA for test samples in Figure \ref{fig:freqF1}.
Firstly, we see the differences are positive, which shows that micro F1 consistently improves across all bins.
Secondly, as expected, the most dramatic improvements are achieved for the codes with the fewest training samples,
and the least improvement is achieved by the region for which we have the highest number of training examples. This was the region which had the highest F1 (Fig. \ref{fig:F1}) score and hence less room for improvement.

\section{Conclusion}
In this paper we introduced MVC-RLDA, a  model for medical code predictions, composed of a stack of embeddings, multi-view CNNs with cross channel max pooling shared across all codes, and separate spatial attention pooling for each code. This model has the potential to flexibly capture the relationship between different n-grams and codes. We further enhance this model by using the descriptions of the labels in regularizing the attention weights to mitigate the effect of overfitting, especially for classes with few training examples. We also demonstrate the advantage of using other notes aside from the discharge summaries. Our model surpasses the previous state-of-the-art model on the MIMIC III dataset, providing more accurate predictions according to numerous metrics. We also presented a detailed analysis of the results to highlight the contributions of our innovations in the achieved result.

The simplest among these was to use all available text in addition to the discharge summary, as our approach was to concatenate all relevant notes in each input.

It is worth exploring more nuanced approaches for integrating other notes in the input, as all notes may not be similarly important. 
Other modifications may yield further improvements.
For instance, we trained the model on all ground-truth codes equally, similarly to previous approaches \cite{baumel2018multi,wang2018joint,mullenbach2018explainable,perotte2013diagnosis}. However, medical codes are ordered according to their importance. It is worth exploring approaches which take the rank of labels into account. Furthermore, devising models which incorporate the hierarchical knowledge of the codes can be helpful. Finally, it will be important to test our model in an AI-assist workflow to see how automated predictions can expedite human coding.
\bibliography{naaclhlt2019}
\bibliographystyle{acl_natbib}

\end{document}